\documentclass[pdflatex,sn-nature]{sn-jnl}

\usepackage{graphicx}%
\usepackage{multirow}%
\usepackage{amsmath,amssymb,amsfonts}%
\usepackage{amsthm}%
\usepackage{mathrsfs}%
\usepackage[title]{appendix}%
\usepackage{xcolor}%
\usepackage{textcomp}%
\usepackage{manyfoot}%
\usepackage{booktabs}%
\usepackage{algorithm}%
\usepackage{algorithmicx}%
\usepackage{algpseudocode}%
\usepackage{listings}%
\usepackage{CJKutf8}%

\begin{document}
\begin{CJK}{UTF8}{gbsn}
\title[Article Title]{Intelligent Chemical Purification Technique Based on Machine Learning}


\author[1]{\fnm{Wenchao} \sur{Wu}}\email{2001111731@stu.pku.edu.cn}
\author[2]{\fnm{Hao} 
\sur{Xu}}\email{2001111740@pku.edu.cn}
\author[2,3]{\fnm{Dongxiao} \sur{Zhang}}\email{dzhang@eitech.edu.cn}
\author*[1,4,5]{\fnm{Fanyang} \sur{Mo}}\email{fmo@pku.edu.cn}

\affil*[1]{\orgdiv{School of Materials Science and Engineering}, \orgname{Peking University}, \city{Beijing}, \postcode{100871}, \country{China}}

\affil[2]{\orgdiv{College of Engineering}, \orgname{Peking University}, \city{Beijing}, \postcode{100871}, \country{China}}

\affil[3]{\orgdiv{Ningbo Institute of Digital Twin}, \orgname{Eastern Institute of Technology}, \orgaddress{\city{Ningbo},  \state{Zhejiang}, \postcode{315200}, \country{China}}}

\affil*[4]{\orgdiv{School of Advanced Materials}, \orgname{Peking University Shenzhen Graduate School}, \orgaddress{\city{Shenzhen}, \postcode{518055}, \country{China}}}

\affil*[5]{\orgdiv{AI for Science (AI4S)-Preferred Program}, \orgname{Peking University Shenzhen Graduate School}, \orgaddress{\city{Shenzhen}, \postcode{518055}, \country{China}}}


\abstract{We present an innovative of artificial intelligence with column chromatography, aiming to resolve inefficiencies and standardize data collection in chemical separation and purification domain. By developing an automated platform for precise data acquisition and employing advanced machine learning algorithms, we constructed predictive models to forecast key separation parameters, thereby enhancing the efficiency and quality of chromatographic processes. The application of transfer learning allows the model to adapt across various column specifications, broadening its utility. A novel metric, separation probability ($S_p$), quantifies the likelihood of effective compound separation, validated through experimental verification. This study signifies a significant step forward int the application of AI in chemical research, offering a scalable solution to traditional chromatography challenges and providing a foundation for future technological advancements in chemical analysis and purification.}




\maketitle

\section{Introduction}\label{sec1}
Column chromatography is a crucial tool\cite{bib1} in modern chemical research, primarily employed for separating and purifying substances. However, traditional column chromatography methods pose significant challenges in terms of efficiency and time, as they often rely on trial-and-error processes and accumulated experience. Even seasoned researchers often spend considerable time experimenting to identify optimal chromatographic conditions, a process typically marked by inefficiency and time consumption.

In recent years, the rapid advancement of Artificial Intelligence (AI) in the field of chemical analysis and process optimization\cite{bib2,bib3,bib4} has opened up new possibilities for improving traditional methods. Machine learning and data-driven modeling have shown great potential in the recognition of complex biological samples\cite{bib5,bib6} and drug development\cite{bib7,bib8}. These technologies are beginning to transform traditional workflows. Chemists have increasingly turned to advanced techniques, including machine learning and deep learning, to uncover patterns and regularities in complex data, leading to new perspectives and breakthroughs in chemical research\cite{bib9,bib10,bib11}. In our group, we utilize machine learning methods in thin-layer chromatography(TLC)\cite{bib12,bib13} and high-performance liquid chromatography(HPLC)\cite{bib14} to predict the relationship between corresponding chemical structures and their chromatographic retention values.

From the perspective of a data-driven research paradigm, machine learning leverages datasets composed of a significant amount of data. However, despite the fact that column chromatography has been developed for many years\cite{bib1}, researchers often view it as a tool without systematically collecting experimental conditions and parameters related to the process. Chromatographic datasets often have a large number of missing values and low standardisation, which can pose a challenge for further statistic research, even if some data is recorded. Although commercially available automated column chromatography systems save researchers time, they have limitations in data collection due to the constraints of their operating mechanisms, which impede the free monitoring and automated processing of data.

To advance the research and application of column chromatography, it is essential to develop an automated platform capable of consistently testing different samples and recording standardized data. This platform will provide the necessary hardware support to build predictive models for column chromatography, potentially overcoming the limitations of traditional methods and improving efficiency. 

Our study aims to develop an effective predictive model for chromatography and provide constructive suggestions. To this end, an automated experimental setup was developed first to collect standardized chromatographic data. Advanced machine learning algorithms were then used in conjunction with extensive experimental data to train and validate the model. The main objective of the research is to develop an AI-based chromatography model for predicting and optimizing key parameters in the separation process, thereby improving separation efficiency and purification quality. The success of this research could bring considerable changes to chemical analysis and purification methods, particularly in the fields of organic synthesis and materials discovery, where it holds the potential to significantly improve research and development(R\&D) efficiency.

\section{Data acquisition}\label{sec2}

After conducting initial investigations and an extensive literature review, it has become evident that establishing an automated column chromatography platform is crucial for the efficient and precise collection of data. Automation not only enhances efficiency but also ensures the stability and quality of data, significantly reducing errors that can arise from human intervention. The high quality and reliability of the collected data lay a solid foundation for thorough data analysis and subsequent modeling. Additionally, the research paradigm combining automation with machine learning technology has already demonstrated its significant advantages in the field of chemical research. This interdisciplinary integration has been successfully applied in various chemical studies, ranging from the optimization of reaction conditions\cite{bib15} to the interpretation of mechanisms\cite{bib16}. Through an automated platform, as the research pipeline shown in the Fig.1a, we can systematically collect a large volume of experimental data, and machine learning technology provides powerful tools to analyze these data, revealing hidden patterns and correlations, thereby deepening our understanding of chemical processes.

\begin{figure}[ht]%
	\centering
	\includegraphics[width=0.9\textwidth]{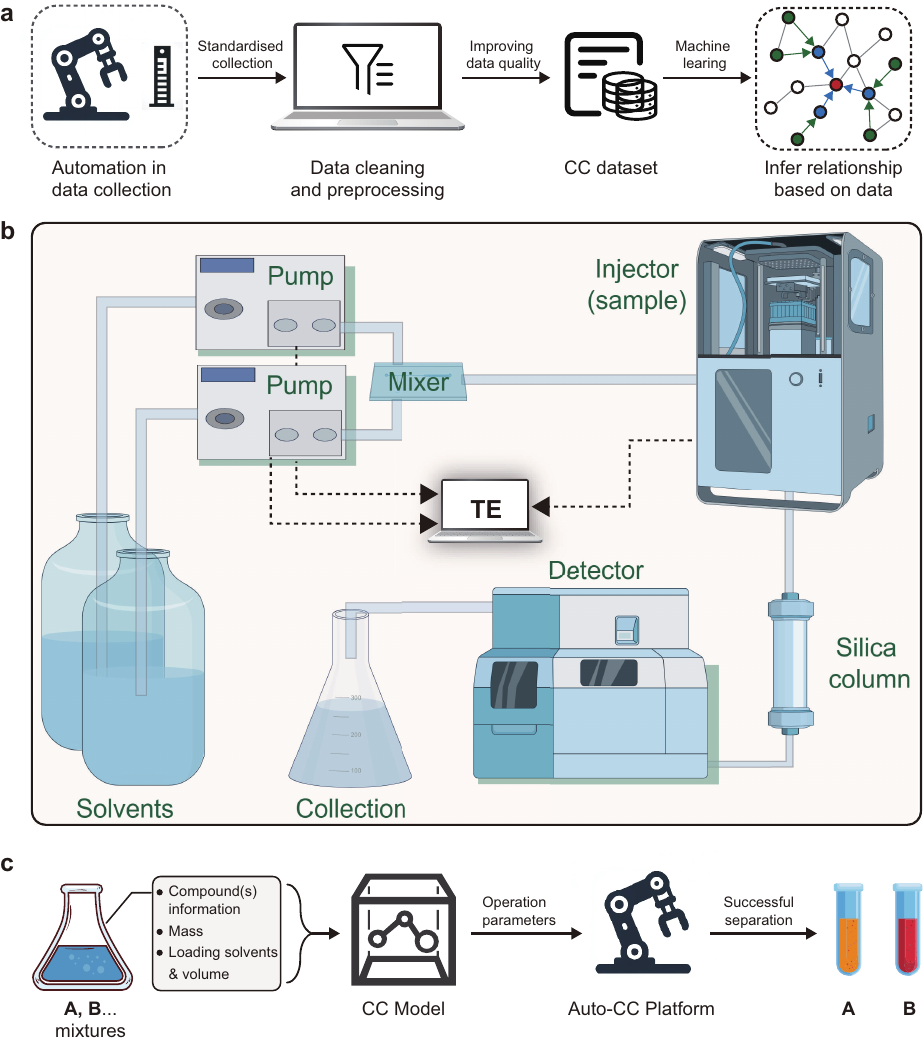}
	\caption{Schematic diagram of main parts. (a) Research pipeline. (b) Schematic of automation platform. (c) Separation process schematic (CC, column chromatography).}\label{fig1}
\end{figure}
In response to the challenges encountered in column chromatography experiments, we have engineered an automated device designed for the efficient collection of data. The structural design and operational mechanism of the device are depicted in Fig.1b, with a photographic representation provided in Supplementary Fig.S1. The device comprises four independent modules: the pumping unit, which is responsible for eluent infusion and consists of two pumps and a mixer; the autosampler, which is responsible for sample loading; the UV detector, which is responsible for detecting chemical signal; and the control terminal, which is responsible for handling these modules for synergistic cooperation. A comprehensive description of the operational procedure can be found in the Method section. This hardware support can break down the barriers to intelligent column chromatography. For example, as shown in Fig.1c, a mixture of compounds \textbf{A} and \textbf{B} is inputted into the prediction model to obtain separation parameters. The model then sends instructions to the experimental platform, resulting in successful separation of the mixture.

In precision separation techniques like column chromatography, the accuracy and reproducibility of results hinge on the meticulous execution of several critical steps. These steps include the precise loading of the sample, the preparation of the mobile phase at a specific ratio, the adjustment of the optimal flow rate (${v}$) for sample separation, and the real-time capture and analysis of compound separation signals to ascertain the start ($t_{1}$) and the end ($t_{2}$) times of the compound's separation. Subsequently, the process proceeds to the next sample in accordance with the established protocol. Automation technology executes every step of the process, from the loading of samples to the calculation of separation volumes, with a level of precision and consistency that surpasses manual operations. This superior performance is facilitated by the meticulous control exerted over each procedural step. The automated control system ensures uniform experimental conditions, which in turn leads to results that are not only highly comparable but also remarkably reliable. Additionally, the use of automation technology boosts the throughput of experiments, enabling a greater number of samples to be analyzed within the same timeframe, thereby significantly enhancing the efficiency of research endeavors.

Nonetheless, the variability in preparative chromatography columns and the differing processes among manufacturers can influence data collection strategies. For example, the Agela Claricep Flash series of preparative chromatography columns, available in common sizes such as 25g and 40g, are characterized by longer lengths and larger internal diameters, packed with a greater volume of material. This can result in experiments taking anywhere from tens of minutes to several hours to complete, which in turn leads to increased solvent consumption and elevated time-related costs. When planning chromatography experiments, it is crucial to take these experimental conditions into account to optimize the process effectively. 

In order to address the aforementioned issue, our investigation focused on employing smaller format columns, such as 4g columns, for data collection. These columns offer the advantages of reduced experimental timeframes and lower solvent usage, which enables the rapid accumulation of a substantial dataset and facilitates the development of a foundational model. We successfully amassed a dataset comprising 4,684 standardized data points for 218 distinct compounds, spanning roughly 20 diverse compound classes, all analysed using the 4g columns. Detailed compound's information can be found in of the Supplementary Table S1. To enhance the generalizability of the experimental outcomes, each compound was subjected to testing across three distinct sample volumes and a range of seven eluent ratios at least. Moreover, we pursued selective data acquisition across columns with diverse specifications and explored the application of transfer learning techniques to accommodate the variability in column characteristics, the forthcoming sections will delve into the specific applications and the effects of these methodologies with a detailed discussion.

\section{Basic model with machine learning techniques}\label{sec3}

Following the completion of the data collection phase on the automated column chromatography platform, we visualized the entire dataset, as depicted in Fig.2a and Fig.2b. This visualization process clarified the structure of the dataset, which was compiled using four different specifications of chromatography columns, with data evenly distributed across the seven eluent ratios. For the reasons previously discussed, it was necessary to attempt the establishment of a basic model based on the 4g dataset. Recognizing that the separation volume is a critical parameter in column chromatography, we conducted an in-depth analysis of the UV spectral curves accurately determining it. This analysis allowed us to pinpoint the separation times for each compound—$t_{1}$, marking the initial detection of the compound, and $t_{2}$, indicating the moment when the compound was fully eluted out from the column. Fig.2c also gives a figurative definition of separation volume in this study, where $V_{1}$ and $V_{2}$ correspond to the volume of eluent consumed at the start of compound separation and at the time of complete elution, respectively. Thus, we have:

\begin{equation}
V_1 =\textit{v} \times t_1\label{}
\end{equation}
\begin{equation}
V_2 =\textit{v} \times t_2\label{}
\end{equation}
\begin{equation}
{\Delta}V=V_2 - V_1 =\textit{v} \times (t_2 - t_1)\label{}
\end{equation}

Before constructing the machine leaning model, an appropriate feature engineering is needed to capture all the important factors that need to be considered in the experiment. Mechanistically, the main factors influencing the $V$ value are molecular polarity and experimental conditions. Molecular polarity is directly related to a compound's structure and properties. The Molecular Access System (MACCS) keys are widely utilized structural keys that encapsulate information about the molecular structure, its fragments, and substructures. These keys allow for the transformation of any compound into a 167-dimensional representation. Additionally, the RDKit software facilitates the straightforward extraction sixteen kinds of molecular properties (detailed in Method section), including molecular weight (MolWt) and the number of hydrogen bond donors (HBD), etc. Moreover, there are several key parameters for column chromatography experiments that need to be integrated, including the proportion of eluents, the type of loading eluent (e), compound mass (m), and the volume of loading eluent ($V_{e}$), converting these into algorithmic terms is a set of 9-dimensional experimental parameter features, consisting of 6-dimensional weighted descriptors for solvents and 3-dimensional column chromatography column information. In a systematic investigation to ascertain the predictive accuracy of diverse machine learning strategies, we scrutinized three distinct algorithms: Light Gradient Boosting Machine (LGB), Artificial Neural Networks (ANN), and Graph Neural Networks (GNN). Various algorithms handle parameters differently. For traditional algorithms like LGB and ANN, parameters such as the feature engineering shown in Fig. 2c are inputted directly into the training function without any special handling.

In our previous study\cite{bib14}, we explored a geometry-enhanced graph neural network approach to molecular representation. This approach utilizes the natural graphical properties of molecular structures and integrates problem-relevant features into the graph representation. It combines the 3D conformation of the molecule, experimental conditions, relevant descriptors, and quantile learning techniques\cite{bib17} to meet the specific needs of the chemical domain. The use of the quantile learning technique is essential to capture the uncertainty in the training process, providing a reasonable confidence interval for the predicted results. Additionally, combining expertise in mathematical modelling of chromatographic processes and column properties with machine learning algorithms further enhances the predictive accuracy of the models. This integrated approach has improved the efficiency and quality of data collection. It also provides a solid foundation for predictive modelling in column chromatography. This is expected to lead to significant advances in the field of chemical analysis and purification.  

The QGeoGNN model employed in our study adopts a succinct and efficient strategy for feature input. As illustrated in Fig.2d, the 3D conformation of a molecule can be divided into a Graph \textit{G} describing atom-bond and a Graph \textit{H} describing bond-angle. Fig.2e provides a detailed illustration of the network architecture formation. The model's feature input comprises 9-dimensional experimental parameter features and 16-dimensional molecular features, which are embedded into two distinct graph representations. In concrete terms, the 9-dimensional experimental parameter features are embedded into the edge features of Graph \textit{G}, representing experimental data. These edge features reflect the various proportions of eluents used in the chromatography experiments and other key experimental parameters, providing specific information about the experimental conditions to the model. Afterwards, the 16-dimensional molecular features are embedded into the edge features of Graph \textit{H}. These molecular features, which include chemical and physical properties of the molecules, are crucial for capturing the interaction between molecules and silica particles, essential for predicting their behavior in column chromatography. The model then utilizes Graph Isomorphism Convolution (GIC) techniques\cite{bib18} to generate specific graph representations based on these embedded node and edge features. GIC is a powerful graph neural network operation capable of effectively capturing complex relationships and patterns between nodes in a graph. In the case of GNN, the column information is structured in a comparable manner and integrated into Graph \textit{G}. Further implementation details of these models are outlined in the Method section.
 
  \begin{figure}[H]%
	\centering
	\includegraphics[width=0.8\textwidth]{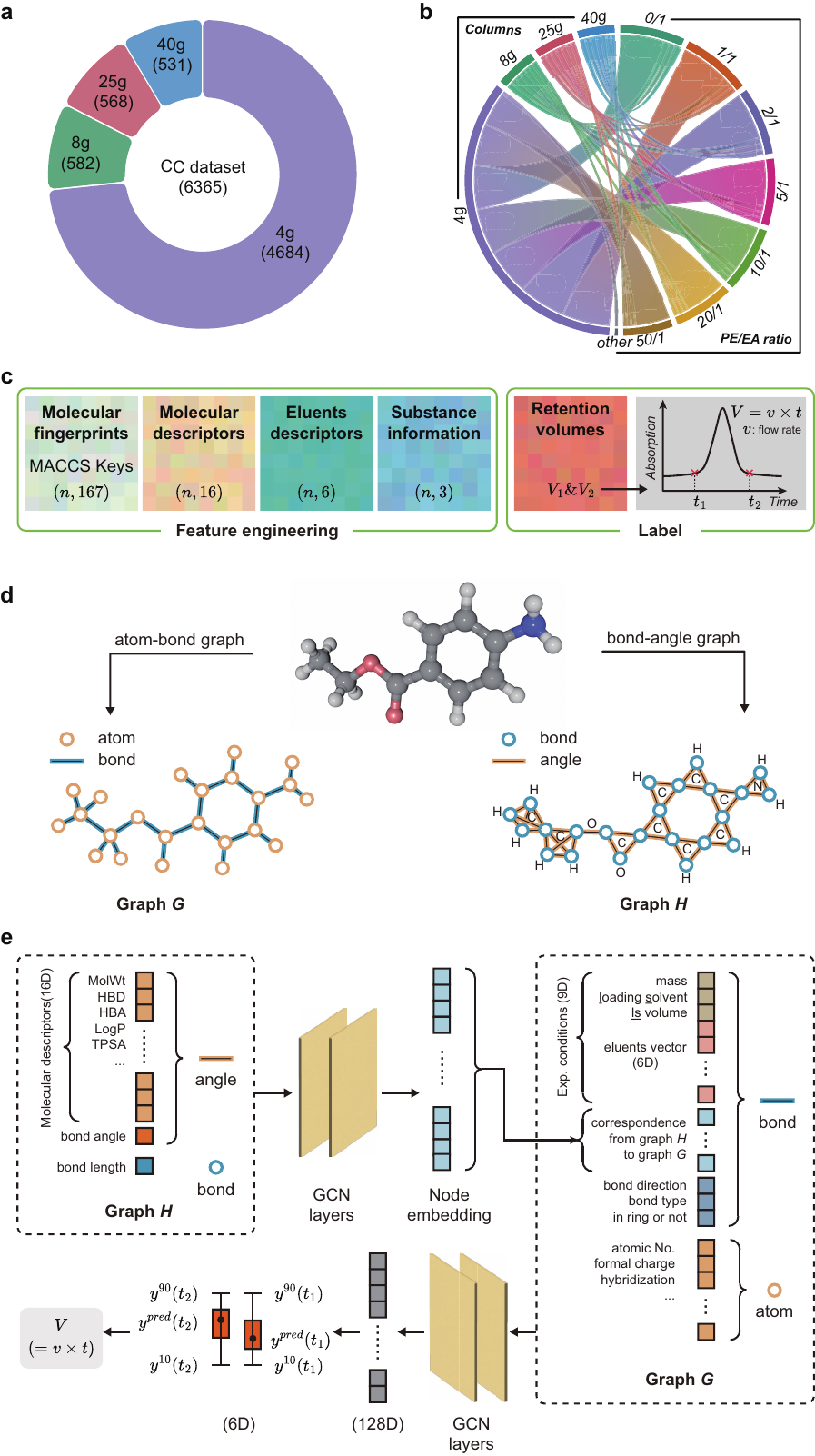}
	\caption{Dataset distribution and feature engineering. (a) Column specifications for number distribution. (b) Distribution of the amount of data according to the proportion of eluents. (c) Feature engineering for ANN and LGB. (d) The schematic instruction of atom-bond Graph \textit{G} and bond-angle Graph \textit{H}. (e) Schematic diagram of QGeoGNN hierarchy.}\label{fig2}
  \end{figure}

These models, each with its inherent advantages and idiosyncrasies, were evaluated to furnish a holistic view of the efficacies of varied algorithms in processing chemical datasets. For this analysis, the dataset pertaining to column chromatography was judiciously partitioned into three subsets: 80$\%$ was dedicated to the training set, with the remainder equally divided between validation and test sets (10$\%$ each). This stratagem was intended to allocate ample data for model training while reserving independent subsets for validation and testing, thereby facilitating a robust assessment of the models’ generalization capacities and empirical predictive prowess. The training set underpinned model learning and parameter optimization, the validation set served for model selection and hyperparameter refinement, and the test set was utilized for the ultimate evaluation of predictive performance. This methodological approach enabled a rigorous comparison of the performances of diverse machine learning algorithms on a uniform dataset, laying a scientific foundation for the selection of the most apt model for predicting outcomes in column chromatography. Such a step is indispensable for guaranteeing the model’s reliability and precision, particularly when navigating the complexities inherent in chemical data analysis, where disparate machine learning models may demonstrate markedly different performance profiles.

\begin{figure}[ht]%
	\centering
	\includegraphics[width=0.9\textwidth]{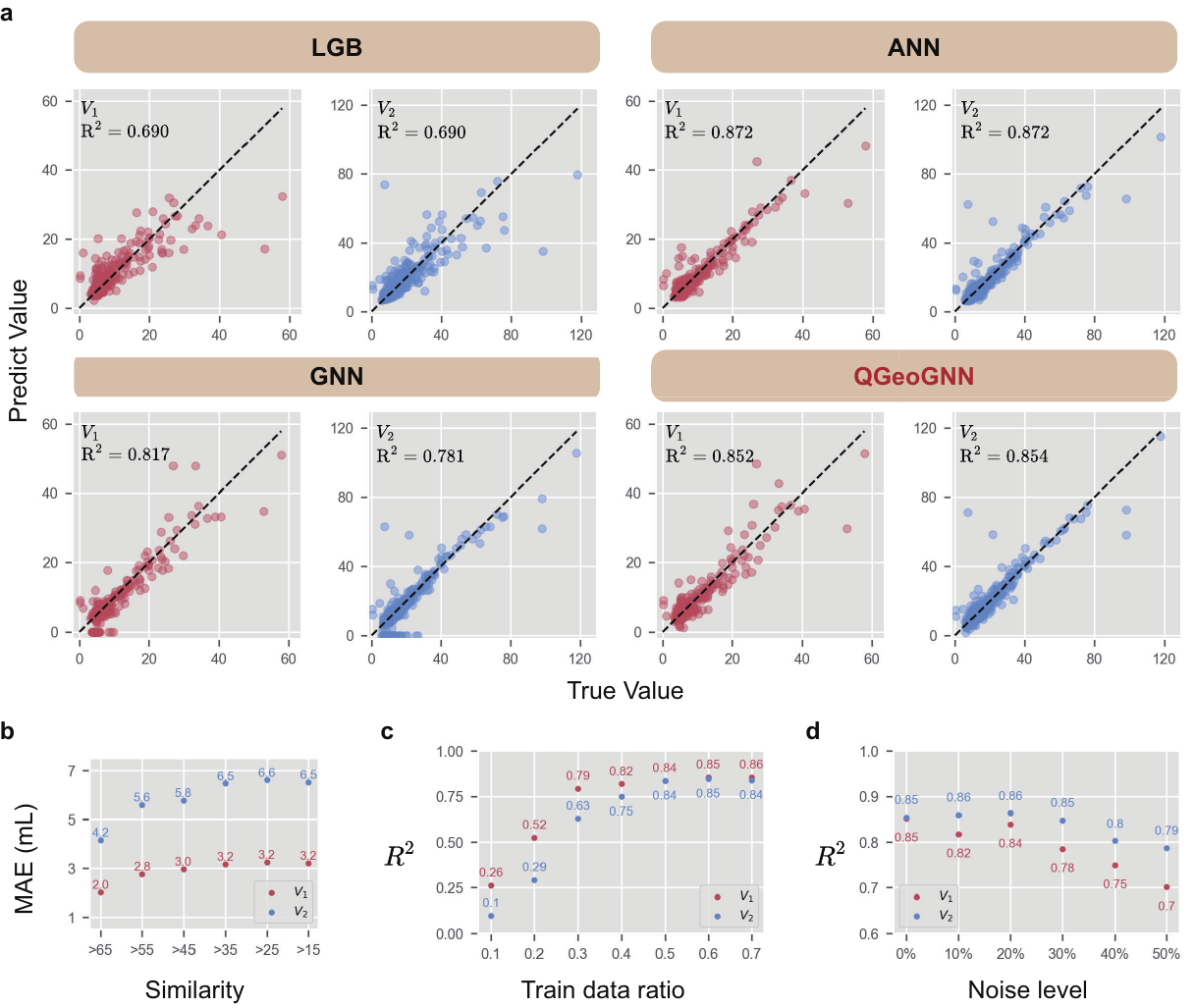}
	\caption{The training results of the basic model. (a) Training outcomes of different machine learning algorithms based on 4g-dataset. (b) Relationship between mean absolute error (MAE) and compound similarity. (c) Impact of the training set proportion on \(R^{2}\). (d) Influence of Gaussian noise ratio on $R^{2}$.}\label{fig3}
\end{figure}

The study's model learning was based on data from a 4g chromatography column, with the results shown in Fig.3a. These results indicate that the LGB, a tree-based model, has somewhat limited predictive accuracy, with $R^{2}$ values for $V_1$ and $V_2$ being only around 0.69. In contrast, ANN and QGeoGNN demonstrate closer $R^{2}$ performances. Drawing from prior work in the HPLC domain, it's noted that QGeoGNN is capable of integrating both experimental parameters and molecular features effectively. Importantly, as the volume of data increases, the predictive performance of QGeoGNN shows an upward trend, which is crucial for handling large-scale chromatography data. Furthermore, it is observed that traditional GNN, which only include a single graph representation (Graph \textit{G}), perform relatively poorly in terms of $R^{2}$.

Based on the analysis and findings, the decision to utilize the QGeoGNN to construct a predictive model for column chromatography is well-founded. This choice is predicated on the advantages of QGeoGNN in handling complex chemical data, as well as its potential for enhanced performance in a big data environment. The implementation of QGeoGNN is expected to make the column chromatography prediction model more promising, offering more accurate predictions for compound separation. Such advancements are likely to play a significant role in the application of these models in synthetic laboratories, paving the way for more efficient and precise experimental processes.

To comprehensively assess the performance of our developed predictive model, a series of tests were conducted. These tests were designed to reveal the model's behavior under various conditions, including the relationship between compound similarity and model prediction error (i.e., MAE), as well as the impact of training set proportion and noise ratio on model prediction accuracy ($R^{2}$). Initially, we explored the relationship between compound similarity and MAE through 20-fold cross-validation (Fig.3b). The results indicate that higher compound similarity corresponds to lower error in predicting separation volume, aligning with the general expectation that models predict chromatographic behavior more accurately when the structural similarity of compounds is high. Experimental data demonstrated that the $R^{2}$ value tends to stabilize when the training set proportion exceeds 50$\%$ and the noise ratio is maintained below 20$\%$ (Fig.3c,d). This suggests that under these conditions of dataset size and noise level, the model achieves robust predictive performance, indicating its resilience to training data quantity and noise. Furthermore, validation experiments were conducted to determine if different compounds mixed together have the same retention volume as they do when in pure form. The experiments were performed on a few compounds that separated under specified conditions. The results (Fig.S2) demonstrated that when the compounds have no interaction with each other and are within the appropriate mass range, their retention volume remains the same when mixed together as it does when they are in pure form.

\section{Transfer learning}\label{sec4}
In the context of this study, acknowledging the constraints imposed by time and solvent expenses, the volume of data accrued from chromatography columns of specifications other than the 4g benchmark was approximately 10$\%$ of the latter's dataset. Consequently, we implemented a parameter transfer strategy\cite{bib19,bib20,bib21} to refine the column chromatography model, with the aim of accommodating chromatography columns of varied specifications whilst facilitating the training of larger models on limited datasets. The quintessence of transfer learning lies in its capacity to leverage knowledge acquired from one task to bolster learning efficiency and performance on another, closely related task. In this vein, specific segments of the extant model were meticulously adjusted to more aptly align with the requirements of the new dataset, thus epitomizing the strategic application of transfer learning principles in our research.

In the input layer of the model, we integrated information pertinent to new chromatographic column specifications (Fig.4a). This integration is a critical step as it enables the model to accommodate and process the unique characteristics of new chromatographic columns, ensuring the model's ability to recognize and adapt to these novel operational conditions. The output layer of the model underwent updates to account for variations that different column specifications might introduce to the final results. Consequently, adjustments to the model's final stages were necessary to accurately predict outputs for new specifications. This approach allows the model to adapt to new output requirements while retaining its existing knowledge base.

During the application of transfer learning, we opted for a strategy that involved transferring the parameters of the original network to a new network and adopted a lower learning rate ($10^{-4}$) during the training part. Compared to the data volume for the 4g column, these smaller-scale tasks could lead to overfitting\cite{bib22,bib23} when training complex models like QGeoGNN from scratch. This decision was primarily based on three considerations: Firstly, a reduced learning rate helps to prevent excessive modification of the learned features in the original model\cite{bib25}, which has been pretrained on a substantial amount of data and thus possesses a good generalization ability for certain features. A lower learning rate ensures that these fine-grained feature representations are not compromised by rapid weight updates. Secondly, excessively high learning rates might lead the model to overfit specific samples in the new dataset, overlooking the generalization ability of the original knowledge\cite{bib26}. Finally, given the relatively smaller scale of the new dataset, a lower learning rate helps to mitigate the risk of overfitting. Additionally, we leveraged features learned from the 4g base model, such as three-dimensional molecular information, to expedite training and conserve computational resources and time for the new models\cite{bib24}. By gradually adjusting the model weights, we can more effectively balance the knowledge between the new and the old datasets, thereby enhancing the model's generalization performance in new chromatographic column specifications.

\begin{figure}[ht]%
	\centering
	\includegraphics[width=0.9\textwidth]{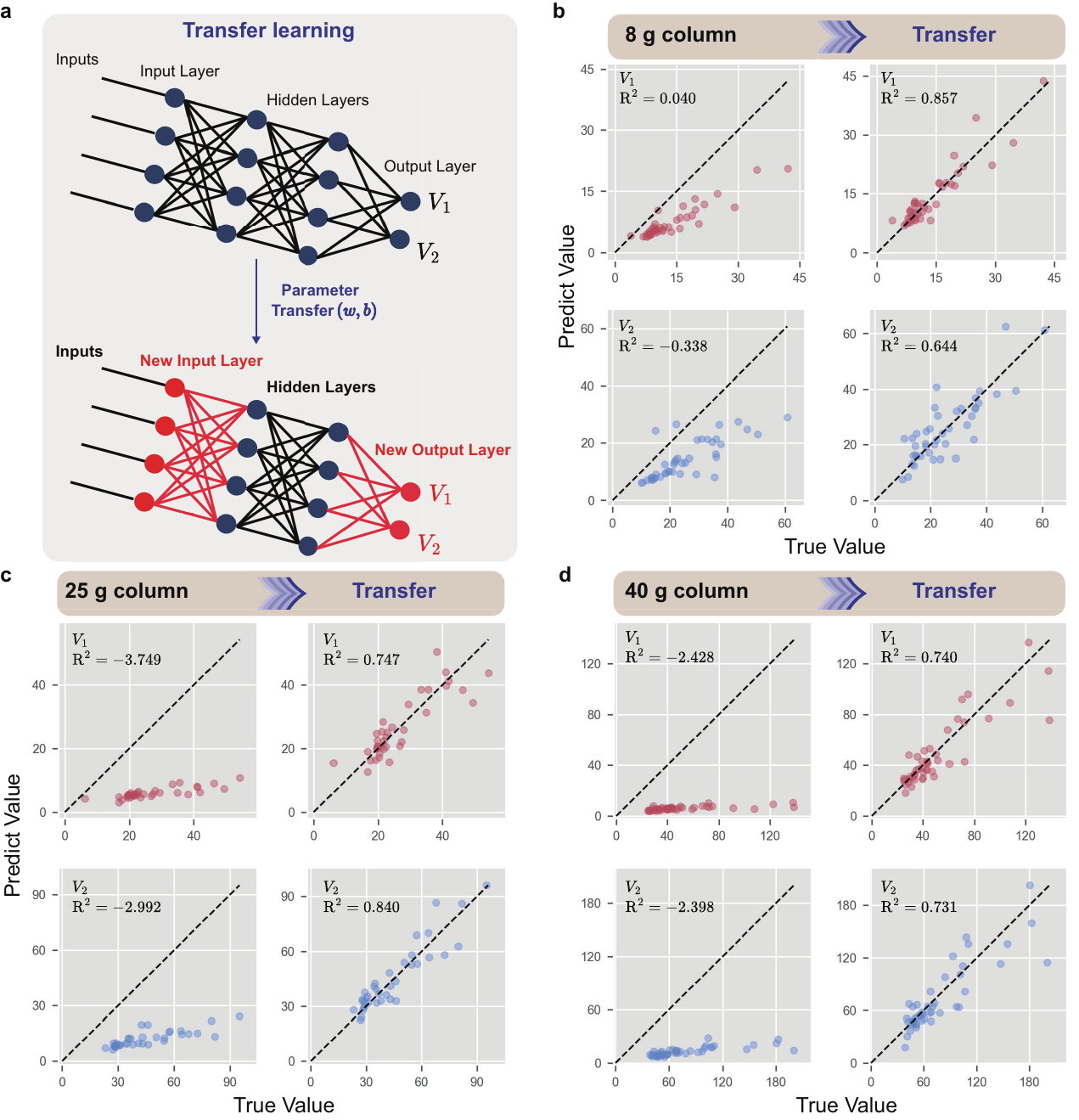}
	\caption{Transfer learning and training results. (a) Transfer learning process; (b, c, d) Training results from the application of transfer learning to the 8g, 25g, and 40g column datasets using the QGeoGNN model initially developed for the 4g.}\label{fig4}
\end{figure}

Operationally, we selected the basic model with the minimum MAE on the 4g column, updated its output layer, and fine-tuned these parts using parameters from the new tasks. The application of transfer learning methods to chromatography column datasets of different specifications (8g, 25g, and 40g) and the comparison of their effectiveness against direct training methods reveal significant insights. Specifically, for the 8g column (Fig.4b), we initially employed the same direct training approach used for the 4g column. However, this approach yielded lower prediction accuracy, with $R^2$ values for $V_1$ and $V_2$ being only 0.04 and -0.338, respectively. This outcome indicates that the direct training method is not sufficiently effective when dealing with new chromatography column specifications. In contrast, the application of the transfer learning method resulted in a significant improvement in the model's predictive capabilities. Specifically, on the 8g chromatography column dataset, the model adjusted through transfer learning achieved $R^2$ values of 0.857 and 0.644 for $V_1$ and $V_2$, respectively. Similarly, we observed a comparable trend in datasets for the 25g and 40g chromatography columns (Fig.4c, d). This substantial improvement not only validates the effectiveness of transfer learning in enhancing model predictive capability, but also highlights its suitability for handling data from chromatography columns of varying specifications.

\section{\texorpdfstring{Application: $S_p$}{Application: Sp}}\label{sec5}

To quantitatively assess the separation probability of compounds under specific conditions, we introduced a separation probability formula based on quantiles. This formula calculates the separation probability based on the quantiles of elution volumes ($V$) of two compounds during the chromatographic process. The calculation method for the $S_p$ involves specific quantile-based measurements. We define the 90th percentile of the $V_2$ of the compound \textbf{A} as $V_{A2}^{90}$ and the 10th percentile of the $V_1$ of the compound \textbf{B} as $V_{B1}^{10}$. Similarly, the 90th percentile of the $V_1$ of the compound \textbf{B} is $V_{B1}^{90}$, and the 10th percentile of the $V_2$ of the compound \textbf{A} is $V_{A2}^{10}$. As depicted in Fig.5a, assuming that compound \textbf{A} be eluented out first, the $S_{p}$  is then derived based on these quantile values:

\begin{equation}
S_{p}=1-\frac{\max \left(0, V_{A2}^{90}-V_{B1}^{10}\right)}{V_{B1}^{90}-V_{A2}^{10}}\label{}
\end{equation}

The logic of this formula is based on the overlap of the elution times of the two compounds in chromatography column. If the value of ($V_{A2}^{90}-V_{B1}^{10}$) is positive, it suggests that $V_{B1}$ could be less than $V_{A2}$, indicating that the separation of the two compounds may not be fully achieved. If the value is negative, it indicates complete separation of the two compounds, with a separation probability of 1.
\begin{figure}[ht]%
	\centering
	\includegraphics[width=0.9\textwidth]{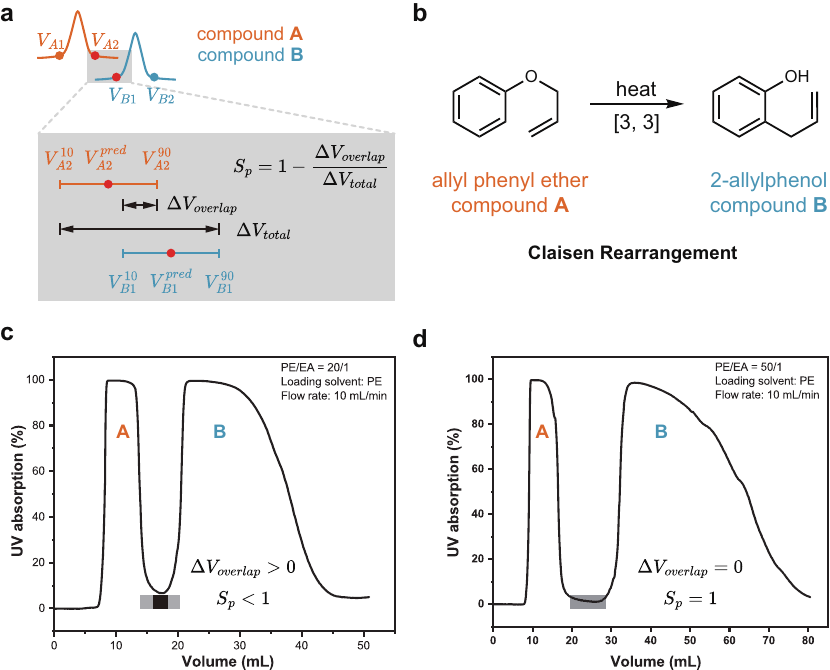}
	\caption{Application of $S_p$ in CC predictive model. (a) Schematic diagram of $S_p$ calculation. (b) The classic Claisen rearrangement reaction. (c, d) Wet laboratory validation based on model prediction results. }\label{fig5}
\end{figure}

While our model demonstrated high predictive accuracy within the dataset, we recognize that accuracy alone is insufficient for comprehensively addressing column chromatography challenges. Meanwhile, more important is separating the chemical products outside the dataset. To validate the model's generalizability, we selected a classic chemical reaction, the Claisen reaction\cite{bib28}, as the experimental subject (Fig.5b). In this reaction, due to its reversibility, both reactant and product coexist in the mixture after reaction, necessitating the separation of high-purity products. Initially, we utilized the predictive model to determine the parameters required for effective separation. In the validation experiment, we may wish to mix pure allyl phenyl ester (compound \textbf{A}) and 2-allylphenol (compound \textbf{B}) directly as an alternative to the reaction mixture. Thus, they were mixed in a 1:1 ratio and added to a sufficient quantity of sample solvent to meet the minimum sample volume requirements of the sampler. Subsequently, the experiment was initiated at the control terminal, yielding the corresponding experimental results as shown in Fig.5c and Fig.5d. From the corresponding separation probability is obtained by quantile calculation, when PE/EA = 20/1, the predicted value of $S_{p}$ is 0.51. It is clear that there is indeed some intersection in the separation process of A and B, that means they're not completely separate. And when PE/EA = 50/1,  the predicted value of $S_{p}$ is 1.0, there is no intersection in the separation process of A and B, that means they're completely separated. This validation result demonstrates the practical value of $S_{p}$ in guiding the separation of actual products. Using the Claisen reaction as an example, our model accurately predicted outcomes under different experimental conditions. Detailed prediction data can be found in Supplementary Table S2.

\section{Discussion}\label{sec}
This research developed machine learing models for column chromatography, addressing the challenges of inaccurate and reproducibility issues of subjective experiences that chemists traditionally face in their experimental procedures. We constructed an automated chromatography experimental platform for standardized data collection. We then employed the latest machine learning methods, selecting the QGeoGNN algorithm to build the basic model. This model was subsequently adapted for chromatography columns of various specifications through transfer learning techniques. QGeoGNN, integrating molecular three-dimensional conformations, experimental conditions, relevant descriptors, and quantile learning techniques, was effectively tailored to the specific needs of the chemical domain. Additionally, by combining column characteristics with machine learning technology, we further enhanced the predictive performance of the model. We also defined a separation probability metric to measure the likelihood of compound separation under specific conditions.
 
Nonetheless, this study faces limitations that require further work. First, the dataset's compound quantity is relatively limited compared to the broader range of compounds, leading to missing categories and potential impacts on model performance. Second, the eluent system used here is PE/EA, which is generally suitable for less polar compounds. For compounds that are difficult to separate in this system, other eluent systems such as methanol/chloroform, petroleum ether/acetone, etc. may be used. However, due to the lack of data for these systems, the scope of the model will be limited accordingly. Lastly, the definition of $S_{p}$ in this study is based on the objective laws of column chromatography experiments, it is important to acknowledge that chromatographic conditions in practice can be influenced by various factors, including column temperature and the nature of the stationary phase, etc. These factors have not been adequately accounted for in current models and may have a practical impact on the separation results. Nevertheless, our research demonstrated that the data-driven approach employed in this study is effective to column chromatography retention volumn prediction. Further research should investigate the impact of these variables on the model's predictive ability and attempt to integrate them to enhance its adaptability and accuracy. With these enhancements, we can anticipate the model providing more dependable separation predictions across a broader range of application scenarios. Plans are in place to use feedback from the broader chemistry community on the open-source model for empirical validation and optimization, aiming to enhance the model's accuracy and utility.

\section*{Method}
\subsection*{Construction and control of automation equipment}
To address the challenges encountered in column chromatography experiments, we have designed and assembled an advanced automated apparatus specifically for the collection of column chromatography data. The construction and operational principle of the device are depicted in Fig.1b (a photograph of the apparatus is provided in Fig.S1). The system incorporates two medium-pressure pumps (manufactured by Beijing Qingbohua Technology Co., Ltd.), each delivering different pure solvents (petroleum ether and ethyl acetate) with a flow accuracy of 0.01 mL/min and a maximum flow capacity of 200 mL/min. To facilitate the mixing of eluents in multiple proportions, the outlets of these pumps are connected to a mixer, ensuring that the mobile phase can be thoroughly mixed in predetermined ratios before the elution step. The centerpiece of the system is an autosampler, which stores and automatically transports the prepared sample solutions, activating automatically upon receiving programmed instructions. Real-time signals during the separation process are captured and recorded by an ultraviolet detector operating at a wavelength of 254 nm. For real-time response and data synchronization, the pumps, detector, and autosampler are all connected via RS232 serial ports for real-time communication. Control over the entire automated platform is executed through a Python program, which manages sample positioning, flow rate adjustments, eluent proportioning, and the collection, conversion, analysis, and archiving of real-time detection signals. A video of the experiment is provided in the Supplementary information.

\subsection*{Construction of the CC dataset}
In this study, a column chromatography dataset comprising separation volume data for 218 compounds was collected using an automated experimental platform. The dataset encompasses 6365 records of separation volumes, along with experimental conditions and column information. The dataset utilized for pre-training the basic model consists of 4684 entries of 218 compounds, employing columns of 4g specifications and includes comparisons of two different flow rates. The datasets for transfer learning correspond to different column specifications, with 582 data points of 90 compounds using 8g columns, 568 data points of 80 compounds using 25g columns, and 531 data points of 81 compounds using 40g columns. Notably, the compounds for different columns are all within the scope of the 218 compounds, with detailed distributions available in the Supplementary information.

Flow rates were set based on the manufacturer's recommendations: 10 mL/min for a 4g column, 10 mL/min for an 8g column consisting of two 4g columns connected in series, 15 mL/min for a 25g column, and 30 mL/min for a 40g column. These settings were found to be reasonable and efficient during pre-tests, resulting in high-quality data collection. The photos of the chromatographic columns are also displayed in Fig.S3. It took six months to collect the data, using approximately 600 L of petroleum ether, 180 L of ethyl acetate, and 20 L of dichloromethane (DCM). The DCM is utilised to load samples and clean the injection syringe.

All data indices are compiled in an Excel spreadsheet, where compounds are indexed by their CAS numbers, and column specifications are differentiated in the column specification field. Recorded variables include flow rate, eluent ratio, compound purity, compound density, sample loading mass/volume, solvent volume for sample loading, with the `ending time' denoting the name of the UV original data. These identifiers are used to index corresponding data for subsequent analysis. Upon analysis completion, two columns, \(t_1\) and \(t_2\), are generated, representing the start and end times of separation, respectively. A value of -1 for \(t_1\) and \(t_2\) indicates invalid data. The SMILES descriptors for compounds are generated using a Python script that converts molecular names into SMILES notation.

\subsection*{The graph representation for molecules}
In this study, molecular representation was achieved by constructing two graphs, Graph \textit{G} and Graph \textit{H}, as illustrated in Fig.2d. Graphs are frequently employed in data science to depict unstructured data such as social networks, chemical molecules, and transportation networks. A typical graph consists of multiple nodes and edges that represent the connection relationships between them\cite{bib14}. As shown in Fig.2e, Graph \textit{G} represents the planar structure of a chemical molecule, with nodes corresponding to atoms and edges representing chemical bonds. Each node in Graph \textit{G} encompasses nine attributes of the corresponding atom, including atomic number, chiral tag, degree, explicit valence, formal charge, hybridization, implicit valence, aromaticity, and the number of connected hydrogen atoms. The features of each edge in Graph \textit{G} include three properties of the bond and four experimental condition attributes; the bond properties comprise bond direction, bond type, and whether it is part of a ring, while the four experimental condition attributes include eluent ratio, sample mass, type of sample solvent, and mass of sample solvent. On the other hand, Graph \textit{H} describes the three-dimensional conformation of the molecule, where nodes represent bonds and edges represent bond angles. Each node in Graph \textit{H} contains only one attribute, the length of the corresponding bond, whereas each edge features 17 properties, including the bond angle and 16 related descriptors such as molecular weight (MolWt), number of hydrogen bond donors (HBD), number of hydrogen bond acceptors (HBA), the octanol/water partition coefficient (LogP), total polar surface area (TPSA), and 11 molecular descriptors calculated by the Python package Mordred (listed in Supplementary Table S3), selected based on the Spearman coefficient which identifies the correlation with separation time.

\subsection*{Strategies for transfer learning}
To elucidate the generality and adaptability of our model across various chemical datasets, the current research adopts a strategy of transfer learning. In the regime of transfer learning, the model is primed with weights acquired from antecedent training endeavors. This approach is eminently suited for contexts with constrained data availability or inherent similarities across tasks, thereby augmenting the model’s proficiency in assimilating data from novel domains. Training of the model engaged the mean squared error (MSE) as the criterion for loss, appraising the divergence between predicted outcomes and actual observations. A learning rate scheduler (StepLR) was employed to dynamically modulate the learning rate, enhancing the training protocol, with the ultimate learning rate established at 0.0001. Through relentless performance evaluations on the validation dataset, the generalization capacity of the model was meticulously monitored, culminating in a definitive performance appraisal on the test dataset to validate the model's effectiveness and precision.

\subsection*{Experimental verification}
In this study, several similarity thresholds are utilized to differentiate the similarity level including 65$\%$, 55$\%$, 45$\%$, 35$\%$, 25$\%$, and 15$\%$. Therefore, all data are categorized into six groups where the molecules in each group as the testing dataset have at least one similar molecule in the training dataset above a specific similarity threshold. Our objective was to assess the impact of training set size and data noise ratio on the performance of machine learning models. To this end, we systematically varied the training set proportions and noise ratios, setting them at 0.1, 0.2, 0.3, 0.4, 0.5, 0.6, 0.7 and 0$\%$, 10$\%$, 20$\%$, 30$\%$, 40$\%$, 50$\%$ respectively. Through this approach, we explored the robustness and generalization capabilities of the model when faced with varying degrees of data noise and different magnitudes of training data. Data preprocessing, including cleaning, normalization, and feature selection, was undertaken to ensure data quality. The dataset was partitioned according to the specified proportions of the training set to evaluate the influence of different volumes of training data on model performance. In each experiment, a certain proportion of the data was reserved as a test set to assess the model's generalization ability.

\subsection*{Derivation of chromatographic separation probability}
In this work, we have defined a measurement method termed as chromatographic separation probability, \(S_p\), to quantify the likelihood of a machine learning model correctly separating the mixture of compounds under specific experimental conditions. This section elaborates on the definition and derivation of chromatographic separation probability. Utilizing the proposed QGeoGNN, we can ascertain the range of separated volume for the compounds. The definition of \(S_p\) is grounded on a straightforward principle: regions within the numerical range overlap are deemed inseparable, whereas others are considered separable. Thus, \(S_p\) can be articulated as equation (\ref{sec4}). To derive the formula for calculating \(S_p\), for the scenario where the numerical ranges of compounds partially overlap, as depicted in Fig.5a, the predicted overlap region is considered inseparable, while other areas are separable. Consequently, the overlapping and total parts can be calculated as follows:

\begin{equation}
{\Delta}V_{overlap}=V_{A2}^{90}-V_{B1}^{10}\label{}
\end{equation}
\begin{equation}
{\Delta}V_{total}=V_{B1}^{90}-V_{A2}^{10}\label{}
\end{equation}
the \(S_p\) can be derived as:
\begin{equation}
S_{p}=1-\frac{{\Delta}V_{overlap}}{{\Delta}V_{total}}=1-\frac{V_{A2}^{90}-V_{B1}^{10}}{V_{B1}^{90}-V_{A2}^{10}}
\end{equation}
Here, $V_{A2}^{90}$ and $V_{A2}^{10}$ are the maximum of 90th percentiles and minimum of 10th percentiles for \(V_2\) of the compound \textbf{A}, $V_{B1}^{90}$ and $V_{B1}^{10}$ are the maximum of 90th percentiles and minimum of 10th percentiles for \(V_1\) of the compound \textbf{B}, respectively. It is obvious that the compounds are predicted to be separable within the range of error, and the \(S_p\) = 1. In general, the \(S_p\) can be summarized as:

\begin{equation}
S_{p}=\left\{
\begin{aligned}1-\frac{V_{A2}^{90}-V_{B1}^{10}}{V_{B1}^{90}-V_{A2}^{10}},V_{A2}^{90} > V_{B1}^{10}\\1,\hspace{30pt} V_{A2}^{90} \leq V_{B1}^{10}\end{aligned}\right.
\end{equation}
which can be simplified as:
\begin{equation}
S_{p}=1-\frac{\max \left(0, V_{A2}^{90}-V_{B1}^{10}\right)}{V_{B1}^{90}-V_{A2}^{10}}
\end{equation}

Hence, the defined chromatographic separation probability \(S_p\) spans from 0 to 1. A higher \(S_p\) value indicates an expanded region of separability, in other words, an increased likelihood that the mixture of compounds is predicted to be separable by the QGeoGNN.

\subsection*{Experimental settings and parameters}
In the QGeoGNN utilized in this work, the number of GINConv is 5, the graph pooling strategy is the summation, the embedding dimension of the node and edge representation is 128, and the batch size is 2048. The training epoch is 1500, and the validate loss is adapted for early stopping. For pre-train prediction, the prediction models are established for 4g columns. The optimizer is Adam and the learning rate is 0.001. The subdataset is randomly divided into 80/10/10 to obtain the training, validating, and testing dataset. For comparison, the LGB, ANN and GNN are also employed to train a predictive model. The input of LGB and ANN is composed of the 167-dimensional MACCS keys that are utilized to represent the molecular structure, the 16-dimensional molecular descriptors that are the same as those utilized in QGeoGNN, and 9-dimensional experimental parameter features. For LGB, the maximum depth is 5, the learning rate is 0.007, the number of leaves is 25, and the number of estimators is 1000. For ANN, there are 3 hidden layers with 50 hidden neurons in each hidden layer. The activation function is leaky ReLu and the optimizer is Adam with a learning rate of 0.001. The training epoch is 10,000 and early stopping is adopted. The construction of GNN is similar to QGeoGNN while it only has Graph \textit{G}, and the loss function is of the mean squared error between the predicted and observed value. The experimental parameter features is incorporated into the edge features for GNN. For transfer learing, the individual secondary datasets are split into 80/10/10 to train a transfer model.

\section*{Data availability}
The CC dataset generated in this study have been deposited in GitHub repository.
\section*{Code availability}
All original code can be found in a GitHub repository.
\section*{Acknowledgements}
This work is supported by the Natural Science Foundation of China (Grant Nos. 22071004, 21933001, 22150013, received by F. M.).
\section*{Author contributions}
W.W. and F.M. established the automated platform, W.W. conducted experiments and collected the column chromatography dataset. W.W and H.X. analyzed the data. H.X. performed chemoinformatic and machine learning studies. W.W. and F.M. wrote the manuscript. F.M. conceived the idea and designed the overall research. F.M. and D.Z. supervised the whole project.
\section*{Competing interests}
F.M., W.W., H.X., and D.Z. are inventors on a patent applications (CN 202410282149.7) submitted by Peking University that cover a ML method for CC conditions prediction.
\section*{Supplementary information}
\textbf{Supplementary Information}\\
Supplementary Fig. S1–S3, Discussion and Table S1–S3.\\
\textbf{Supplementary Video}\\
Experimental video of automated column chromatography.\\
\textbf{Supplementary Data}\\
Compounds used in the experiment Excel statistics.








\bibliography{bibliography}

\end{CJK}
\end{document}